# SKG-LLM: Developing a Mathematical Model for Stroke Knowledge Graph Construction Using Large Language Models

**Ali Sarabadani[1], Kheirolah Rahsepar Fard[2]\*, Hamid Dalvand[3]**

[1] *Department of Computer Engineering and Information Technology, University of Qom, Qom, Iran*

[2] *Department of Computer Engineering and Information Technology, University of Qom, Qom, Iran*

[3] *Department of Occupational Therapy, School of Rehabilitation, Tehran University of Medical Sciences, Tehran, Iran*

## *Abstract*

The purpose of this study is to introduce SKG-LLM. A knowledge graph (KG) is constructed from stroke-related articles using mathematical and large language models (LLMs). SKG-LLM extracts and organizes complex relationships from the biomedical literature, using it to increase the accuracy and depth of KG in stroke research. In the proposed method, GPT-4 was used for data pre-processing, and the extraction of embeddings was also done by GPT-4 in the whole KG construction process. The performance of the proposed model was tested with two evaluation criteria: Precision and Recall. For further validation of the proposed model, GPT-4 was used. Compared with Wikidata and WN18RR, the proposed KG-LLM approach performs better, especially in precision and recall. By including GPT-4 in the preprocessing process, the SKG-LLM model achieved a precision score of 0.906 and a recall score of 0.923. Expert reviews further improved the results and increased precision to 0.923 and recall to 0.918. The knowledge graph constructed by SKG-LLM contains 2692 nodes and 5012 edges, which are 13 distinct types of nodes and 24 types of edges.

\* Email address of the Corresponding Author : rahsepar@qom.ac.ir

**Introduction**

Biomedicine is a discipline with a vast amount of highly specialized knowledge recorded from biological experiments and clinical practice. The growing volume of biomedical literature presents opportunities and challenges for researchers seeking to extract valuable insights and knowledge [1].

In the field of biomedical informatics, big data is a new topic that has attracted a lot of research. Broadly, the characteristics of big data are defined by the three main characteristics of volume, variety, and velocity, commonly known as the 3Vs. First and foremost, the volume of data is growing exponentially in biomedical fields. The second characteristic of big data is the variety of data types and structures. The biomedical big data ecosystem includes different levels of data sources to create a rich set of data for researchers. The third characteristic of big data refers to the speed and processing of data, and various technologies of big data are designed according to the speed of data generation[2].

In the past decade, efforts have been made to collect and manage the vast amount of biomedical knowledge. Knowledge graphs (BKGs) have emerged as a new paradigm for better management of large-scale and heterogeneous biomedical knowledge and have recently attracted considerable interest in academic and industrial literature. Stroke is a research area of biology. Understanding the complex relationships between different factors of this disease is very important for the advancement of medical knowledge. Understanding these relationships using traditional knowledge extraction methods from the text is a difficult task, and often, the complex and multidimensional nature cannot be depicted by these methods [3].

In the field of large language models (LLM), a great revolution has occurred in the last decade due to the availability of big data and the advancement of computing technologies [4]. This LLM revolution has paved the way for a series of new models. Increasing the scale of language models has been able to create better performance and better efficiency in a wide range of NLP tasks. Meanwhile, knowledge graphs as structural representations of the formation of relationships between entities In a specific domain, these representational structures are widely used in various applications[5].

We propose the SKG-LLM complex model to address these challenges and construct a comprehensive KG of stroke-related articles. GPT-4 was considered as the basis of work in data preprocessing to create based data embedding, which enables the extraction and accurate representation of complex relationships using advanced probabilistic techniques.

The idea of using GPT-4 and its advanced language processing capabilities enables more accurate identification and disambiguation of entities, and its use can be useful for building KG to extract complex relationships with high accuracy and efficiency. It also leads to the production of comprehensive and reliable graphs that integrate data from different sources seamlessly. The proposed SKG-LLM model proposed in this research integrates the mutual information of two approaches, Bayesian networks and tensor decomposition to model and extract detailed entity interactions. The SKG-LLM approach increases the accuracy of relationship extraction by these two possible approaches, and on the other hand, provides a strong foundation for the analysis of stroke-related data.

Traditional precision and Recall evaluation criteria were used to validate the desired model. GPT-4 was also used for further evaluation. In the continuation of the process of this article, the details of the SKG-LLM approach, including data set extraction, data preprocessing, parameter estimation, entity and relationship extraction, and model optimization are described.

The remainder of this paper will be organized as follows: Section 1 will cover previous research in the area. Section 2 will describe in detail the process and methods we used in our study. Section 3 will describe nodes and edges and the result of visualization, respectively. Section 4 will further investigate the evaluation process, and finally, Section 5 concludes the paper and provides insight.

## 1. Related works

Scientific articles in the field of biomedicine are increasing day by day and increasing the need for data mining and text mining techniques. This literature is mainly available in the form of structured and unstructured texts. The information contained in them is vital for biomedical research and applications. Therefore, we need biomedical literature mining (BLM) techniques[6]. Many efforts and studies have been made on this topic in biomedical informatics (BMI) and computer science (CS). In the following, we discuss some of the most important research conducted in biomedicine and Large Language Models.

In [7], the authors investigated the synergistic potential of LLMs and medical KGs in predicting diagnoses given by electronic health records (EHR) under the framework of Retrieval Augmented Manufacturing (RAG). They proposed a new graph model called DR.KNOWS. DR.KNOWS selects the most relevant pathology knowledge paths based on medical problem descriptions. To evaluate DR.KNOWS, they developed the first comprehensive human evaluation approach to evaluate the performance of LLMs to predict diagnosis and examine the logic behind their decision-making processes, with the aim of improving diagnostic safety. Using real-world hospital datasets, their study helps enrich the discourse on the role of medical KGs. This approach achieved inferable results. In multi-shot settings, with and without DR.KNOWS recovered trajectories, ChatGPT achieved an average diagnostic accuracy of 66% and a significant average score of over 94% in reasoning according to human evaluation.

KG-based LLM reasoning methods have challenges, the most important of which is that these models only consider KGs as factual knowledge bases and ignore the importance of their structural information for reasoning. For this purpose, the authors of [8] proposed a new method called reasoning on graphs (RoG), which synergizes LLMs with KGs to provide more interpretable reasoning. In this approach, the RoG first establishes the communication paths established by the KGs as loyalty programs. This framework is then used to retrieve valid reasoning paths from KGs so that LLMs can perform faithful reasoning. Furthermore, RoG not only extracts knowledge from KGs to improve the reasoning ability of LLMs through training but also enables seamless integration with any arbitrary LLM during inference. Extensive experiments on two KGQA benchmark datasets show that RoG performs better on KG reasoning tasks and produces faithful and interpretable results. From the reported results, we can see that the performance of all LLMs is significantly improved by integrating the RoG planning module. Hits of ChatGP on Alpaca=8.5%, LLaMA2=15.3%, and Flan-T5=119.3%. These results show that the RoG planning module can be seamlessly integrated with other LLMs to improve performance without retraining. In [9], the authors presented a new and practical pipeline for constructing a heart failure knowledge graph. This pipeline uses large language models and modification by medical experts. Classical methods based on BERT have a fundamental weakness because they require a large amount of training data to ensure the model's performance. On the other hand, real-world medical annotation data, especially disease-specific annotation examples, are minimal. In addition, BERT models do not perform well in out-of-distribution relationships that are not trained in the training phase. In this study, the authors applied rapid engineering in schema design, information extraction, and knowledge completion stages. Among all the examined results, the best performance was achieved by designing task-specific notification patterns and the TwoStepChat approach. In addition, their method saves 65% of the time compared to manual annotation and is more suitable for extracting information outside the real-world distribution. The TwoStepChat model in BioRED data achieved precision = 83.50, recall = 80.45, and F1 = 81.96 in the NER task. Also, this model achieved Precision=68.25, Recall=67.67, and F1=67.96 in the RE task.

[10] suggested AsdKB to quickly gain knowledge about autism spectrum disorder and help with screening as well as for its early diagnosis. BERT was used to create this knowledge base. AsdKB is a Chinese knowledge base about autism spectrum disorder. The structure of this knowledge base based on different sources is as follows:

1. Disease knowledge from clinical descriptions of SNOMED CT and ICD-10 regarding mental and behavioral disorders.
2. Diagnostic knowledge of DSM-5 and various screening tools recommended by social organizations and medical institutions
3. Expert knowledge about professional doctors and hospitals from the web

AsdKB contains ontological and actual knowledge edges and is accessible as linked data at https://w3id.org/asdkb/. AsdKB has various uses, including answering questions, assisting with diagnosis, and expert advice. Table 1 gives some of the most important text-sci-LLM (Text-Sci-LLM) models in Clinical Modeling.

Evaluation of KGs using LLMs has become an attractive focus of study. There have been several works in the literature that show how knowledge graphs can enhance language models. Guan et al. [21] used folk knowledge graphs to generate data for fine-tuning GPT-2 to increase the language model's ability to generate coherent and non-repetitive stories. In another study, Xu et al. [55] used a knowledge graph to fine-tune language models for story generation. When needed, they added a mechanism to extract relevant information from the knowledge graph based on predicted keywords.

Table 1. most important text-sci-LLM (Text-Sci-LLM) models in Clinical Modeling

| Model | References | Year | Code |
| --- | --- | --- | --- |
| ClinicalBERT | [11] | 2019 | https://github.com/kexinhuang12345/clinicalBERT |
| MEDITRON-70B | [12] | 2023 | https://github.com/epflLLM/meditron |
| ClinicalGPT | [13] | 2023 | - |
| Qilin-Med | [14] | 2023 | https://github.com/williamliujl/Qilin-Med/tree/master |
| ChatDoctor | [15] | 2023 | https://github.com/Kent0n-Li/ChatDoctor |
| HuaTuo | [16] | 2023 | https://github.com/SCIR-HI/Huatuo-Llama-Med-Chinese |
| Baize | [17] | 2023 | https://github.com/project-baize/baize-chatbot |
| Medical mT5 | [18] | 2024 | - |
| Me LLaMA | [19] | 2024 | https://github.com/BIDS-Xu-Lab/Me-LLaMA |
| BiMediX | [20] | 2014 | https://github.com/mbzuai-oryx/BiMediX |

Guu et al. [23] used an external knowledge base to train a language model in answering open-domain questions. In the meantime, studies have been conducted on dynamic knowledge graphs in conversation state tracking programs to model conversation participants [24][25][26][27]. Wang et al. [28] used a statistical model to predict entity relationships from domain-specific filtered text. Nayak and Ng [29] focused on extracting overlapping relationships, where entities have multiple relationships reflected in a text span. All these works aim to extract knowledge about the world from text and create an extensive database that is generally applicable or always describes a specific domain.

## 2. Methodology

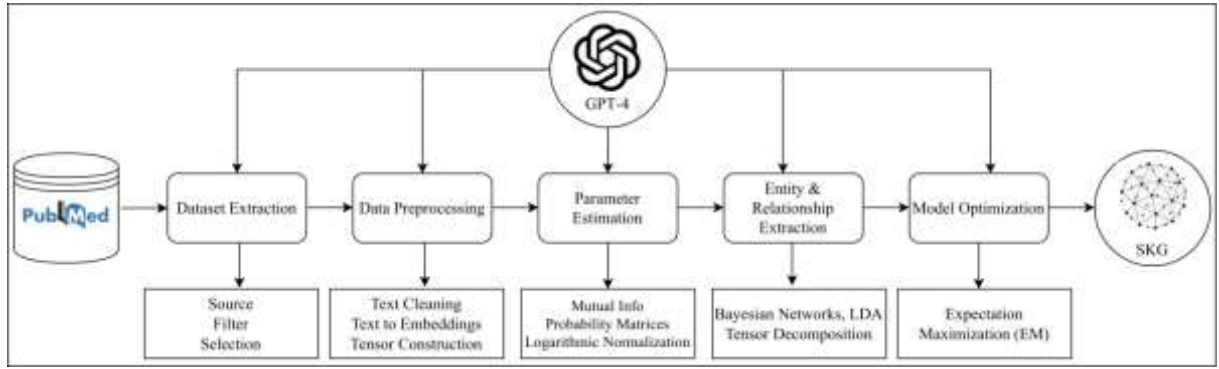

**Figure-1** An overview of SKG-LLM.

Figure 1 shows an overview of the proposed SKG-LLM construction process. In general, the proposed SKG-LLM model has the following basic steps:
- Data set extraction: This step collects scientific articles related to strokes from scientific databases.
- Data pre-processing with LLM: The main pre-processing of the model is in this step. These pre-processing tasks include:
1. Clean text using GPT-4
2. Text normalization
3. Creation of embedding vectors
4. Construction of tensors
   - Parameter estimation: This step also includes several sub-steps:
     1. Calculation of mutual information
     2. Construction of probability matrices
     3. Applying logarithmic normalization, which is used to capture relationships accurately
   - Entity and Relationship Extraction: The steps of this phase are as follows:
     1. Using Bayesian networks
     2. Applying Latent Dirichlet Allocation (LDA)
     3. Tensor decomposition, which is used for modeling and extracting relationships
   - Model optimization: According to the continuous space of the model parameters, advanced techniques are used to modify the model parameters in this step.
   - 

*2.1. Dataset Extraction*

Creating a strong and rich data set for KG is one of the challenges of this study. For this purpose, we focused on extracting titles of publications related to stroke research. For this purpose, we focused on the PubMed database, which includes MEDLINE and the extensive biomedical and biomedical resources. The period considered for the collection of papers was

٥

considered between the years 2020 and July 2024. For this purpose, we focused on a set of keywords including: "stroke", "ischemic stroke", "stroke rehabilitation", and "brain" to ensure relevance. Filtering by these keywords and advanced features of PubMed resulted in the selection of practical papers. At the end of the article selection process, a manual review was done to remove irrelevant papers. After these processes, a data set of 1286 papers was collected. These papers formed the basis of the construction of SKG-LLM. Among these papers, 550 papers on "stroke", 400 papers on "ischemic stroke", 220 papers on "stroke rehabilitation", and 116 papers on "brain" were collected.

*2.2. Data Preprocessing with LLMs*

This section includes the steps of using LLMs as pre-processing. Next, we will clean and normalize the text, convert the text to Embeddings, and create data tensors.

*2.2.1. Text Cleaning and Normalization*

The general process of cleaning and normalizing the raw texts of stroke-related articles is at this stage. This process includes removing noises such as unnecessary punctuation marks, numbers, special characters and non-alphabetic characters. The next step is to remove the extra space. After these steps, several processes are applied to normalize the raw text. These processes include converting all characters to lowercase letters, correcting spelling and typographical errors, and finally removing stop words such as "and", "to" and, "in" which have little effect on semantic analysis. The result of this process is a clean and normalized text, which we use the following relationship for simplicity:

$$Cleaned\ Text = LLM(raw\ text) \qquad (1)$$

This pre-processing process is critical for the construction of KG and ensures text consistency and the absence of noise in the data. Cleaned text is an essential input for embedding models and significantly contributes to the creation of high-quality language models.

*2.2.2. Text to Embeddings*

Raw texts must be transformed into meaningful vectors for inputs to learning models. GPT-4 was used as a language model to extract these meaningful vectors, which are also called embedded vectors. High-dimensional embedded vectors are extracted using this model. These extracted vectors can capture the complex semantic features of the text. In general, this process can be shown as follows:

$$Ei = LLM - Embedding(Cleaned\ Text) \qquad (2)$$

*2.2.3. Data Tensors Construction*

The multidimensional tensor ($Tdata$) is constructed to encapsulate the different dimensions of the data in this step. To create this tensor, embedded vectors are used in the previous step. This tensor includes the following three main parts:

1. $E\ entities$ extracted from the text such as: diseases, symptoms, and treatments, which we call $E\ entities$.
2. Relationships between entities such as: causes, treatment, and related to them, which we call $E\ actions$.

3. Additional features related to actions such as intensity and frequency, which we call $E\ attributes$.

At the end, the final tensor is obtained from the combination of these features as follows:

$$T\ data\ = E\ entities\ \otimes E\ actions\ \otimes E\ attributes \qquad (3)$$

*2.3. Parameter Estimation*

In this section, the parameter estimation steps are described in detail. This section has sub-sections, each of which is further detailed.

*2.3.1. Mutual Information Calculation*

The term mutual information here refers to the information obtained about an entity through the presence of another entity. In this step, GPT-4 estimates the parameters by calculating the mutual information between the entities extracted from the text. Mutual information I (X; Y) between two entities, X and Y, is calculated as follows:

$$I(X;Y) = \sum_{x \in X} \sum_{y \in Y} p(x,y) \log\left(\frac{p(x,y)}{p(x)p(y)}\right) \qquad (4)$$

Where p(x,y) is the joint probability distribution of X and Y, and in this equation, the p(x) and p(y) are the marginal probabilities of X and Y.

The values obtained by mutual information are used to show how much knowledge of one entity reduces uncertainty about another. GPT-4 plays an important role by efficiently analyzing large volumes of text to determine these probabilities with high accuracy, and its ability to understand language and text relationships enables it to identify events and connections within the text, which accurate estimates of joint and marginal probabilities are essential.

*2.3.2. Probability Matrices Construction*

The probability matrix shows the different relationships between different entities, which we use the mutual information values calculated in the previous step to build. These matrices are obtained from the high-dimensional tensors created in the previous step and are constructed as follows:

$$P_{entitics} = \begin{bmatrix} p(E_1,E_1) & p(E_1,E_2) & \cdots & p(E_1,E_N) \\ p(E_2,E_1) & p(E_2,E_2) & \cdots & p(E_2,E_N) \\ \vdots & \vdots & \ddots & \vdots \\ p(E_N,E_1) & p(E_N,E_2) & \cdots & p(E_N,E_N) \end{bmatrix} \qquad (5)$$

Where $p(Ei,Ej)$ in this matrix represents the joint probability of entity $Ei$ being related to entity . GPT-4 plays a vital role in this process. This advanced language model automatically and efficiently calculates these probabilities by carefully analyzing the occurrences and patterns within the text.

*2.3.3. Logarithmic Normalization*

A logarithmic transformation is applied to enhance the interpretability and scale of the probability matrices. This transformation helps normalize the values and make them more suitable for further analysis. The transformed probability matrix is obtained as follows:

$$P'_{entities} = \log(P_{entities} + \epsilon) \qquad (6)$$

$\epsilon$ is a small constant that is added to the input of the problem to avoid taking the logarithm of zero. This transformation scales the probability values logarithmically and leads to easier analysis and interpretation of relationships.

GPT-4 was used for logarithmic normalization. This logarithmic transformation compresses the range of probability values and transforms multiplicative relationships into additive relationships.

*2.4. Entity and Relationship Extraction*

Named Entity Recognition (NER) and Relation Extraction (RE) are two processes by which a knowledge graph is formed from text. The first of these steps is NER, which involves the identification and classification of key entities within unstructured text such as names of people, organizations, places, and other important terms or notions.

Within knowledge graph construction, such entities are then counted as nodes that form up the entire graph. For example, in a medical domain, NER would be the extraction of entities like disease, symptoms, or therapies from within medical literature. After all necessary entities are identified, Relation Extraction (RE) identifies and classifies relations among these entities.

It delineates how two or more entities are connected by predicates like "treats", "causes", and "is associated with". RE connects these nodes (entities) into a knowledge graph with a number of edges, which represent the relationships. For example, RE would reveal that a drug treats a disease, which is included in the graph. Combined NER and RE convert unstructured text into a knowledge graph by identifying and structuring respective entities and connecting relationships between those entities.

*2.4.1. Bayesian Networks and LDA*

Advanced probabilistic models extracted the entities and their bit relations from the text. These approaches include Bayesian Networks (BN) and Latent Dirichlet Allocation (LDA), which help probabilistically model relationships between entities. The probability in BN between two nodes is calculated as follows:

$$P(R \mid E_1, E_2) = \frac{\exp\left(\sum_{k=1}^{K} \lambda_k f_k(E_1, E_2, R)\right)}{\sum_{R'} \exp\left(\sum_{k=1}^{K} \lambda_k f_k(E_1, E_2, R')\right)} \quad (7)$$

Where $E1$ and $E2$ are entities, $R$ is the relationship between $E1$ and $E2$, $\lambda k$ are the weights learned for each feature, and $K$ is the total number of features defined for each pair of entities and relationships.

This formula helps to determine the probability between two entities. GPT-4 enhances the process by efficiently identifying entities and suggesting potential relationships based on its deep linguistic understanding.

The LDA (Latent Dirichlet Allocation) model assumes that each document is a mixture of different topics, where each topic is a distribution over words [30]. LDA is used to model topics within the text and identify hidden relationships between entities. The LDA calculation process is represented by the following equation [31]:

$$P(w \mid z) = \frac{\beta_{wz} + \eta}{\sum_{w'} (\beta_{w'z} + \eta)} \quad (8)$$

$$P(z \mid d) = \frac{\theta_{dz} + \alpha}{\sum_{z'} (\theta_{dz'} + \alpha)} \tag{9}$$

Where β and θ are parameters that represent the word-topic and document-topic distributions.

*2.4.2. Tensor Decomposition*

The tensor decomposition technique interacts with factors extracted in previous steps, such as features, actions and entities. In this research, the canonical polyadic decomposition (CP) is used to factorize the data tensor into the sum of the component tensors. This tensor, which we call $Tdata$, is defined as follows:

$$\mathcal{T}_{\text{data}} \approx \sum_{r=1}^{R} \mathbf{a}_r \otimes \mathbf{b}_r \otimes \mathbf{c}_r \tag{10}$$

Here, $R$ is the rank of the decomposition, and $ar$, $br$ and $cr$ are the component vectors for entities, actions, and attributes, respectively. In this equation $\otimes$ Denotes the tensor product.

Using this analysis makes it easier to understand the underlying data. Moreover, we can deal with the interaction between the component vectors by using the decomposed tensors. The probability of the relationship $R$ according to the data tensor $Tdata$ can be estimated as follows:

$$P(R \mid \mathcal{T}_{\text{data}}) \approx \sum_{r=1}^{R} \mathbf{a}_r \mathbf{b}_r \mathbf{c}_r \tag{11}$$

*2.5. Model Optimization*

The optimization phase of the model and model parameters is one of the most important phases of the design, in which the Expectation Maximization (EM) algorithm was used. The parameter space of the model is continuous, which is challenging to find the most optimal ones. EM algorithm is an iterative method to find maximum likelihood estimates of parameters[32].

In the expectation step (E-Step), this algorithm calculates the expected value of the log function according to the conditional distribution of the hidden variables and according to the observed data and the current parameter estimates. This step includes calculating the next probabilities of hidden variables:

$$Q(\Theta \mid \Theta^{(t)}) = \mathbb{E}_{\text{latent}} \left[ \log P(\text{data, latent} \mid \Theta) \mid \text{data}, \Theta^{(t)} \right] \tag{12}$$

Here, $\Theta(t)$ represents the parameter estimates at the $t-$ th iteration, and Q is the auxiliary function that needs to be maximized in the next step.

In the Maximization Step (M-Step), the auxiliary function is maximized with respect to the Θ parameters. This step updates the parameters to increase the likelihood of the observed data given the new parameter estimates:

$$\Theta^{(t+1)} = \arg\max_{\Theta} Q(\Theta \mid \Theta^{(t)}) \tag{13}$$

## 3. Simulation and Results

GPT-4 was used to build the knowledge graph and both types of nodes and edges. This linguistic model extracted entities and edges on the raw texts of stroke articles. This model extracted 13 different nodes (refer to Table 2) and 24 different edges (refer to Table 3). The strength of this model is that it is trained to recognize different biomedical terms and their contexts and allows us to distinguish between different types of nodes.

Table 2. 13 different nodes extracted by GPT-4.

| # | Node Types | Use |
| --- | --- | --- |
| 1 | Diseases | Represents various types of stroke and related conditions (e.g., ischemic stroke, hemorrhagic stroke). |
| 2 | Symptoms | Clinical manifestations associated with stroke (e.g., headache, paralysis). |
| 3 | Risk Factors | Elements that increase the likelihood of stroke (e.g., hypertension, diabetes). |
| 4 | Treatments | Medical interventions and therapies (e.g., tPA, anticoagulants). |
| 5 | Medications | Specific drugs used in stroke treatment (e.g., aspirin, statins). |
| 6 | Procedures | Medical procedures relevant to stroke care (e.g., thrombectomy, carotid endarterectomy). |
| 7 | Genes | Genes associated with stroke risk and pathology (e.g., APOE, MTHFR). |
| 8 | Proteins | Proteins involved in stroke mechanisms (e.g., fibrinogen, C-reactive protein). |
| 9 | Biomarkers | Biological markers indicative of stroke or its severity (e.g., D-dimer, NIHSS score). |
| 10 | Hospitals | Institutions where stroke treatment is administered (e.g., Mayo Clinic, Cleveland Clinic). |
| 11 | Researchers | Individuals conducting stroke research (e.g., neurologists, epidemiologists). |
| 12 | Organizations | Entities involved in stroke research and care (e.g., American Stroke Association, World Health Organization). |
| 13 | Publications | Key articles and studies on stroke (e.g., journal articles, clinical trials). |

Table 3. 24 different edges extracted by GPT-4.

| # | Node Types | Use |
| --- | --- | --- |
| 1 | Causes | Relationships indicating causation (e.g., hypertension causes stroke). |
| 2 | Treats | Indicates treatment relationships (e.g., aspirin treats ischemic stroke). |
| 3 | Associated with | General associations between entities (e.g., diabetes associated with increased stroke risk). |
| 4 | Symptom of | Indicates symptom relationships (e.g., paralysis is a symptom of stroke). |
| 5 | Expressed in | Expression of genes/proteins in conditions (e.g., CRP expressed in stroke patients). |
| 6 | Encoded by | Genes encoding proteins (e.g., APOE encoded by gene). |
| 7 | Biomarker for | Indicates biomarker relationships (e.g., D-dimer is a biomarker for stroke severity). |
| 8 | Occurs in | Geographic or institutional occurrence (e.g., stroke occurs in elderly population). |
| 9 | Diagnosed with | Diagnostic relationships (e.g., patients diagnosed with ischemic stroke). |
| 10 | Develops from | Disease progression relationships (e.g., TIA can develop into stroke). |
| 11 | Has risk factor | Risk factors for diseases (e.g., smoking is a risk factor for stroke). |
| 12 | Prevents | Preventative measures (e.g., statins prevent stroke recurrence). |
| 13 | Monitored by | Monitoring relationships (e.g., stroke severity monitored by NIHSS score). |
| 14 | Published by | Publication relationships (e.g., study published by journal). |
| 15 | Conducted at | Research locations (e.g., clinical trial conducted at hospital). |
| 16 | Funded by | Funding relationships (e.g., research funded by NIH). |
| 17 | Collaborated with | Research collaborations (e.g., researcher collaborated with another researcher). |
| 18 | Regulated by | Regulatory relationships (e.g., protein activity regulated by gene). |
| 19 | Interacts with | Protein-protein interactions (e.g., fibrinogen interacts with other proteins). |
| 20 | Observed in | Observational relationships (e.g., symptom observed in patients). |
| 21 | Studied in | Study relationships (e.g., gene studied in stroke research). |
| 22 | Implemented in | Implementation of treatments (e.g., treatment implemented in clinical practice). |
| 23 | Researched by | Researcher-entity relationships (e.g., stroke researched by neurologist). |
| 24 | Analyzed in | Analytical relationships (e.g., data analyzed in study). |

# 4. Model Evaluation

The performance of the SKG-LLM knowledge graph is evaluated in two ways. In the first method, we evaluate using traditional evaluation criteria. In the second method, the evaluation of LLM includes using large language models such as GPT-4. The model evaluation process is shown in Figure 2.

Table 4- Details related to various knowledge graphs and knowledge graphs of the proposed approach.

| Name | # Node | # Node Types | # Edges | # Edge Types |
|---|---|---|---|---|
| DrKG [33] | 97 K | 13 | 5.8 M | 107 |
| PrimeKG [34] | 129.4 K | 10 | 8.1 M | 30 |
| Gene Ontology [35] | 43 K | 3 | 75 K | 4 |
| GP-KG [36] | 61.1 K | 7 | 124 K | 9 |
| DDKG [39] | 551 | 2 | 2.7 K | 1 |
| Disease Ontology[40] | 11.2 K | 1 | 8.8 K | 2 |
| DrugBank [41] | 7.4 K | 4 | 366 K | 4 |
| PharmKG [42] | 7.6 K | 3 | 500 K | 3 |
| SKG-LLM | 2692 | 13 | 5012 | 24 |

We used several known KGs in the biomedical field to compare the KG extracted by the SKG-LLM approach. Table 4 shows the details of different KGs and the KG of the proposed approach created on 1286 scientific articles on stroke research.

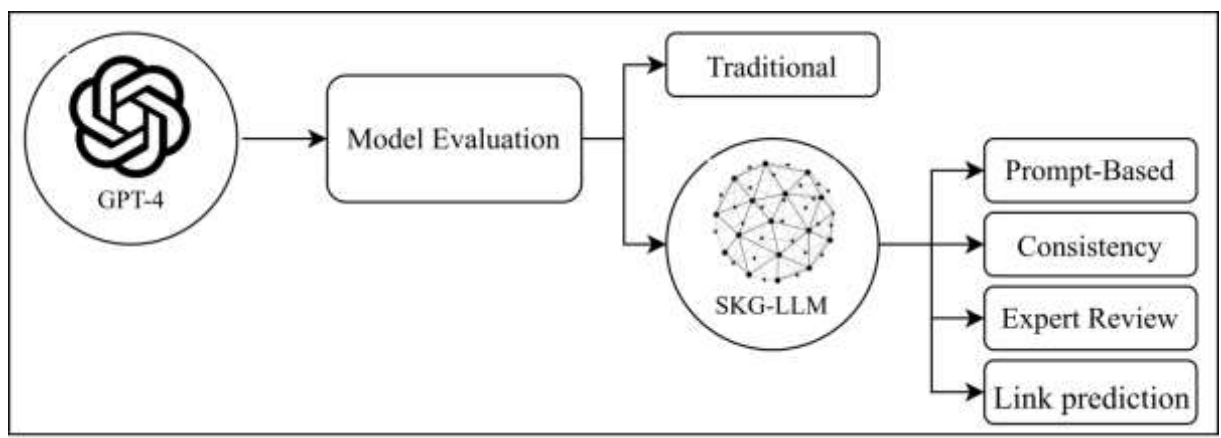

**Figure -2** The process of evaluating the proposed model by traditional approaches and linguistic model- based approach.

*4.1.Traditional Evaluation Methods*

Precision and Recall were used in this evaluation. These two evaluation criteria are defined as follows:

$$\text{Precision} = \frac{TP}{TP+FP} \tag{14}$$

$$\text{Recall} = \frac{TP}{TP+FN} \tag{15}$$

$$\text{F1 Score} = 2 \times \frac{\text{recall} \times \text{precision}}{\text{recall} + \text{precision}} \tag{16}$$

*4.2. LLM Evaluation*

In the LLM evaluation three methods were used: Prompt-Based Validation Consis- tency Check and Expert Examination. these evaluations were practical to check the results were better.

1. Prompt-Based Validation In this assessment speaking is evaluated based on the results of the Representation and the existence of a relationship in GPT-4. In this method 50 targeted prompts were Layouted and Produced for the SKG-LLM Representation. these prompts were specifically organized to value the name rela- tionships betwixt entities. For example questions such as "What is the relationship between high blood pressure and stroke?" or "Is the use of aspirin recommended in the treatment of ischemic stroke?" were posed to the Representation. the Check then responded to these questions exploitation the Removeed information from the articles. Operational Steps:

(a) Layouting the prompts Designing the prompts: Each prompt was specifically designed to assess the relationships between particular entities. For instance, prompts related to diseases, symptoms, treatments, and risk factors.

(b) Check evaluation: the SKG-LLM Check exploitation GPT-4 responded to the prompts.

(c) analyzing the results: The model's responses were compared with known data to measure Precision and Recall.

2. Consistency Check: This Representation was used to describe relationships and confirm the alignment of relationships with the knowledge diagram. inch this wise c important relationships Removeed from the skg-llm Check were evaluated exploitation GPT-4 the end of this rating was to check body and the petit mal epilepsy of contra- dictions betwixt the relationships among different entities inch the cognition graph operational steps:

(a) Selecting relationships: Selecting relationships: From all the extracted relationships, 100 critical relationships were chosen, primarily those involving diseases and treatments.

(b) Consistency evaluation: The selected relationships were compared with other entities and relevant information inch the cognition graphical record to check conjunction and consistency. This evaluation was conducted using GPT-4.

(c) Identifying contradictions: Any inconsistencies or contradictions between the extracted relationships and other entities were identified and flagged for correction.

*4.3. Expert Review*

In this evaluation, experts evaluate the relationships and present the ranking as a consolidated weighted average. For this average, Cohen's kappa was used, which is calculated as follows:

$$\kappa = \frac{p_o - p_e}{1 - p_e} \qquad (17)$$

In this equation po is the observed agreement among raters, and pe is the expected agreement by chance.

Table 5. Result obtained by different model and different evaluation metrics.

| Metric | Wikidata[37] | WN18RR[38] | SKG-LLM |
|---|---|---|---|
| Precision (Traditional) | 0.84 | 0.83 | 0.85 |
| Precision (GPT-4) | 0.89 | 0.88 | 0.87 |
| Precision (Expert Review) | 0.86 | 0.87 | 0.90 |
| Recall (Traditional) | 0.84 | 0.87 | 0.85 |
| Recall (GPT-4) | 0.89 | 0.92 | 0.91 |
| Recall (Expert Review) | 0.87 | 0.88 | 0.88 |
| Consistency (GPT-4) | 0.92 | 0.90 | 0.93 |
| Accuracy (GPT-4) | 0.91 | 0.92 | 0.90 |

Table 5 compares the performance of Relation Extraction (RE) for the three knowledge graphs, namely, Wikidata, WN18RR, and SKG-LLM with respect to metrics like Precision, Recall, Consistency, and Accuracy. In terms of Precision (Traditional), SKG-LLM seems to marginally outperform the other two, claiming a better accurateness in identifying correct relations. But interestingly, when using GPT-4, Wikidata rather holds the highest precision (89.62%), with SKG-LLM a little behind (87.1%). In Expert Review, the highest precision attained by SKG-LLM is 90.7%, indicating that the performance improves indicating a substantial factor as human validation. For Recall (Traditional), WN18RR came first with 87.33, but on employing the services of GPT-4, it captured Recall at 92.76, with SKG-LLM taking the next position at 91. SKG-LLM shows the best Consistency (93.4%) while using GPT-4 meaning that it holds extracted relations in a high degree of coherence within itself. Nonetheless, regarding the Accuracy (GPT-4), Wikidata narrowly wins at 89.66% while SKG-LLM is at the lowest point of 87.5%. This implies that while SKG-LLM performs superiorly in human-reviewed precision and consistency, there is an about area for improvement in accuracy as a whole.

The method of accuracy (GPT-4) is one of the critical evaluation methods in large language models, such as SKG-LLM. Here, the overall accuracy of the model gets evaluated using GPT-4. This essentially means judging the correctness of all relationships drawn from the data, and how aligned they are to scientific facts or existing data. Thus, in other words, Accuracy defines the percentage of all relationships and information created by the model which links with existing and validated data.

Table 6. RE (SKG-LLM , StrokeKG)

| Metric | Precision | Recall | F1 |
|---|---|---|---|
| SKG-LLM (Traditional) | 85.24 | 85.46 | 85.35 |
| SKG-LLM (GPT-4) | 87.11 | 91.03 | 89.02 |
| SKG-LLM (Expert Review) | 90.73 | 88.81 | 89.76 |
| StrokeKG [54] | 80.06 | 88.92 | 84.26 |

It has always been across kinds that SKG-LLM outperforms in expert review scoring highest F1 score of 89.76 (Table 6). This implies that manual validation considerably improves the precision-recall balance. GPT-4 integration also scores on its F1 performance of 89.02, signifying its high automatic extraction capability. On the contrary, StrokeKG records a lower F1 score of 84.26 ostensibly because of its low precision-even though recall is too high. This suggests StrokeKG retrieves relations well, but with higher false positives compared to SKG-LLM.

Table 7- NER(SKG-LLM , StrokeKG, Heart Failure KG )

| Metric | Precision | Recall | F1 |
|---|---|---|---|
| SKG-LLM (Traditional) | 88.62 | 90.24 | 89.42 |
| SKG-LLM (GPT-4) | 92.32 | 89.67 | 90.97 |
| SKG-LLM (Expert Review) | 90.57 | 91.12 | 90.84 |
| StrokeKG | 94.21 | 86.04 | 90.26 |
| Heart Failure KG [55] (TwoStepChat-zeroshot) | 82.33 | 88.50 | 85.31 |
| Heart Failure KG (TwoStepChat-fewshot 10) | 87.35 | 91.35 | 89.31 |

Within the set of comparisons made on the different NER models, SKG-LLM is consistent across all categories, and this is the highest F1 score achieved so far, by SKG-LLM (GPT-4), which is 90.97, thus indicating that integration with GPT-4 increases both precision and recall.  Another very good score is given by SKG-LLM (Expert Review), with an F1 score of 90.84, further proving that expert validation adds value in the maintenance of balanced precision and recall. StrokeKG has a much higher precision (94.21%) but its recall is not good (86.04%) leading to an F1 score of 89.96, which suggests that it can be fairly accurate in identifying entities but missing some relevant ones. The Heart Failure KG models also perform fairly well, where the few-shot (10) model has scored an F1 of 89.31, indicating that few-shot learning does improve the balance between precision and recall compared to zeroshot.

**Definition of Accuracy:** In this case, accuracy is the ratio of correct relationships and information to the total number of relationships and information generated by the model. The formula for accuracy is as follows:

$$\text{Accuracy} = \frac{TP + TN}{TP + TN + FP + FN} \qquad (18)$$

The results of different knowledge graphs and evaluation approaches are shown in Table Table1. In the Precision (Traditional) evaluation criterion, the Wikidata approach achieved precision=0.84, the WN18RR approach achieved precision=0.83, and the proposed SKG-LLM approach achieved precision=0.85, the highest result obtained. In Precision (GPT-4), the Wikidata approach obtained the best result and reached precision=0.89. The proposed approach also obtained the highest result in Precision (Expert Review) and reached precision=0.90, while the Wikidata approach achieved precision=0.86. In Recall (Traditional) and Recall (GPT-4) among the three knowledge nodes, the WN18RR approach obtained the highest value and achieved recall equal to 0.87 and 0.92, respectively. In Recall (Expert Review), two approaches, WN18RR and SKG-LLM, obtained equal results, achieving Recall=0.88. In Consistency (GPT-4), the proposed approach obtained better results and achieved a value of 0.93. Accuracy (GPT-4) in WN18RR was higher than in other approaches.  The summarized diagram of Table 3 is given in Figure 4. This graph shows a comprehensive comparison between -Wikidata, WN18RR, and SKG-LLM - on the investigated metrics. According to this graph, it can be concluded that the SKG-LLM approach has consistently performed better in some criteria.

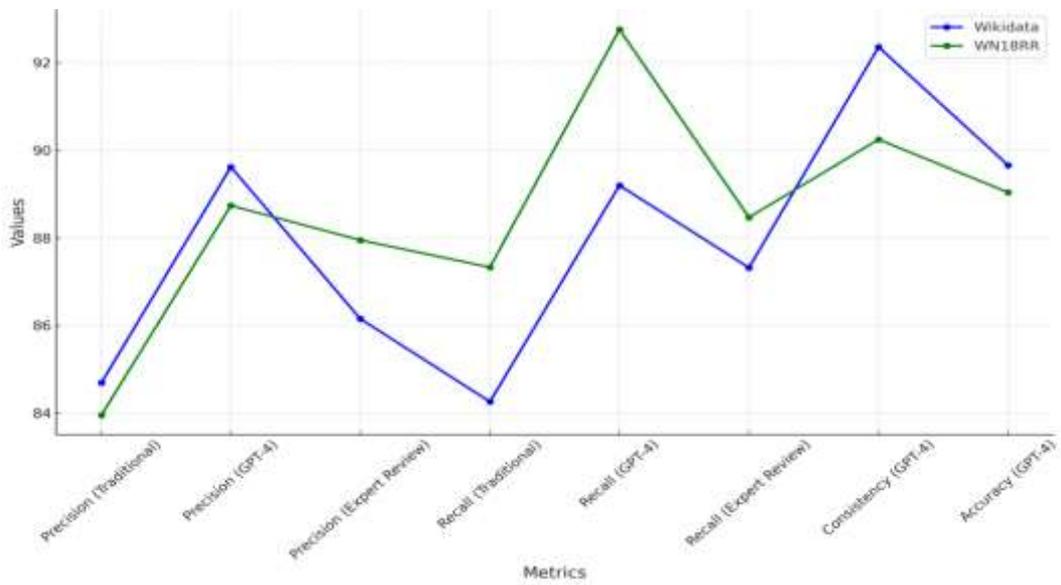

**Figure - 3** Comparison of Wikidata and WN18RR

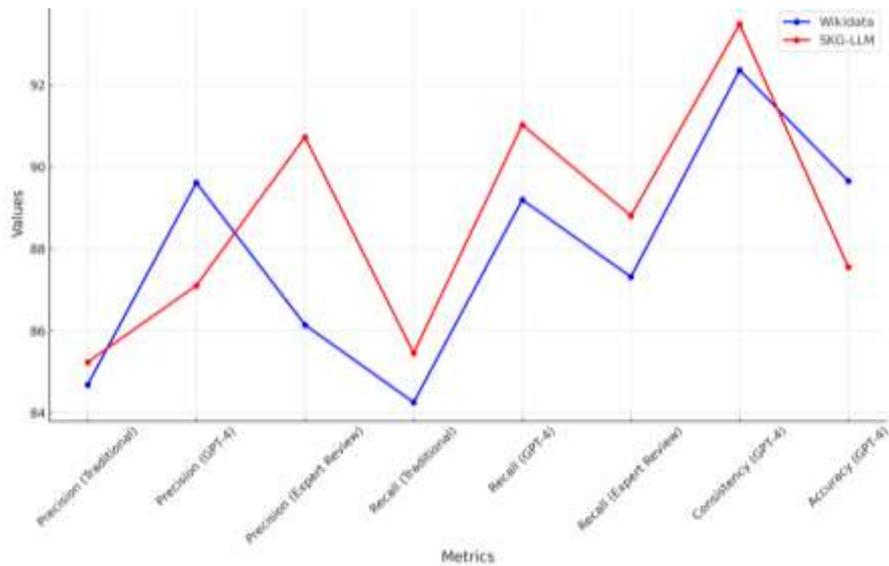

**Figure - 4** Comparison of Wikidata and SKG-LLM

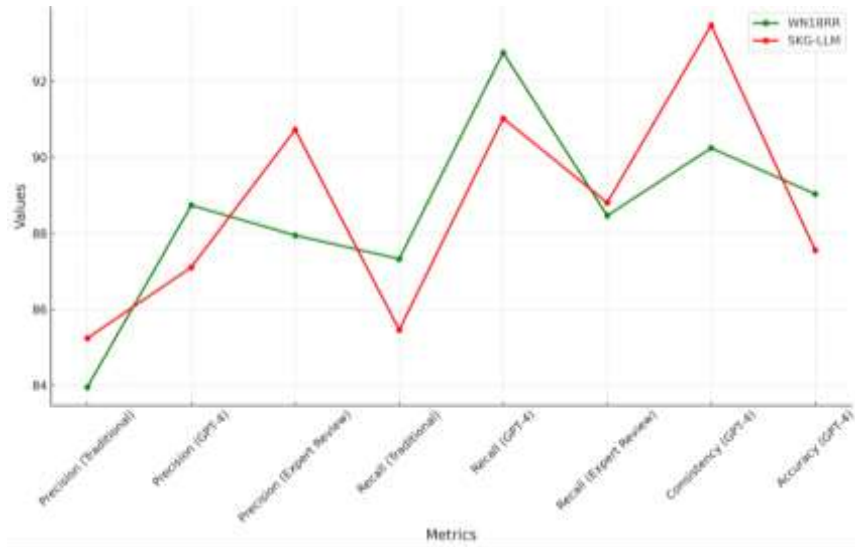

**Figure - 5** Comparison of WN18RR and SKG-LLM

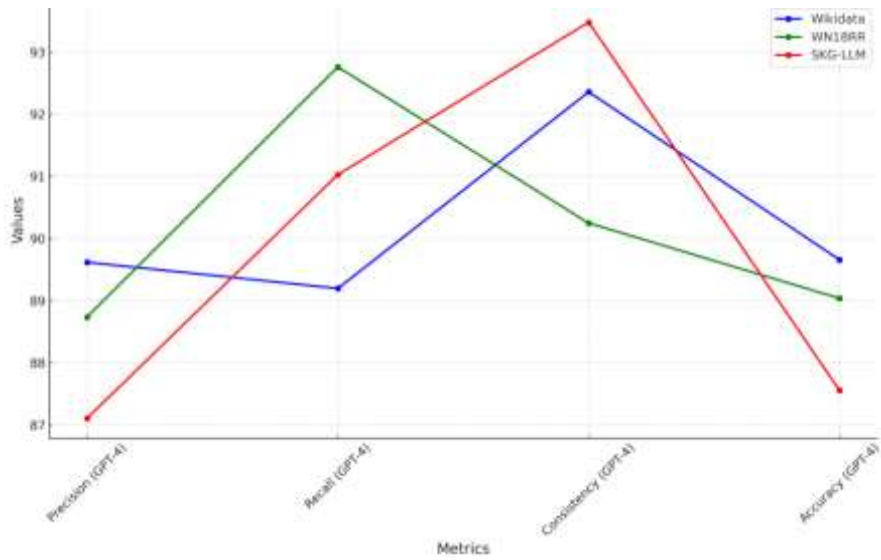

**Figure -6** Comparison of Wikidata ,WN18RR and SKG-LLM for GPT-4 Metrics

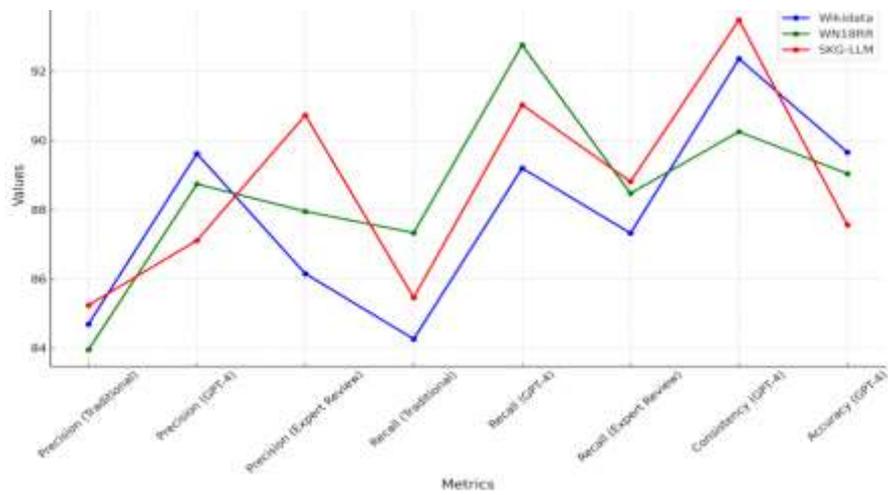

**Figure -7** Comparison of Wikidata ,WN18RR and SKG-LLM Across Evaluation Metrics.

*4.4 Link prediction*

The assessment of CNKG is based on the three metrics, namely, MR, MRR, and P@K, which form the core of our analysis for link prediction. The above are some indicators that provide a fair idea of how well the graph can perform in predicting the types of relationships it establishes between nodes. For this link prediction task, we put the MR, MRR, and P@K in their formulae and consider each separately [39].

1. Mean Rank (MR): MR is a parameter or measure for evaluating the efficacy of information retrieval systems. It measures the average rank of the actual positive items included in a list of items retrieved. This is done by assigning a rank to each item and taking the mean of the ranks of relevant items. A low mean rank indicates better performance as the relevant items are ranked higher on average [39].

$$MR = \frac{1}{N}\sum_{i=1}^{N} rank_i \qquad (19)$$

where $rank_i$ is the rank position of the $i - th$ relevant item, and N is the total number of relevant items.

2. Mean Reciprocal Rank (MRR): Instead of an abstract definition, Mean Reciprocal Rank can be understood well with a practical example. Suppose you search for a specific document in a large databank. MRR is the mean of the reciprocals of the ranks of the first relevant item among the retrieved item list. It is useful in those situations when the first relevant result is most important. MRR gives an idea of how fast the first relevant item is retrieved [40].

$$MRR = \frac{1}{N}\sum_{i=1}^{N} \frac{1}{rank_i} \qquad (20)$$

where $rank_i$ is the rank position of the first relevant item for the $i - th$ query, and N is the total number of queries.

3. Precision at K: P@K is defined as the fraction of relevant documents among the top K retrieved documents. In our analysis, we compute P@K for different values of K: K=1, K=3 and K=10. P@K denotes the relevance of the top K results [40]. It concerns how relevant the top-ranked retrievals are and is calculated as follows:

$$P@K = \frac{1}{N}\sum_{i=1}^{N} \frac{Number\ of\ relevant\ items\ in\ top\ K}{K} \qquad (21)$$

where K is the number of top items considered, and N is the total number of queries.

We utilize link prediction algorithms like TransE, RotatE, DistMult, ComplEx, ConvE, and HolmE to raise some methods that can infer links through advanced algorithms from the graph. The accompanying Table 8 represents the comparison between SKG-LLM and other models in terms of performance metrics such as FB15K-237, WN18RR, and YAGO3-10. Other metrics include Mean Rank (MR), Mean Reciprocal Rank (MRR), Precision at 1 (P@1), Precision at 3 (P@3), and Precision at 10 (P@10). The examination of MRR for SKG-LLM reveals slight superiority or comparable performances in the case of other graphs above all for ComplEx and ConvE on any dataset. For example, SKG-LLM delivers a lower MRR than ComplEx (0.367) and ConvE (0.305) on FB15K-237, where it reaches a score of 0.318, thus slightly lowering its capturing correctness in predictions.

Table 8- Comparison of Link Prediction for SKG-LLM with TransE, RotatE, DistMult, ComplEx, ConvE, and HolmE using MR, MRR, and P@K Metrics with Other KGs.

| KG | | TransE [48] | RotatE [49] | DistMult [50] | ComplEx [51] | ConvE [52] | HolmE[53] |
|---|---|---|---|---|---|---|---|
| FB15k-237 [45] | MR | 209 | 178 | 199 | 144 | 281 | - |
| | MRR | 0.310 | 0.336 | 0.313 | 0.367 | 0.305 | 0.331 |
| | P@1 | 0.217 | 0.238 | 0.224 | 0.271 | 0.219 | 0.237 |
| | P@3 | 0.257 | 0.328 | 0.263 | 0.275 | 0.350 | 0.366 |
| | P@10 | 0.496 | 0.530 | 0.490 | 0.558 | 0.476 | 0.517 |
| WN18RR [46] | MR | 3936 | 3318 | 5913 | 2867 | 4944 | - |
| | MRR | 0.206 | 0.475 | 0.433 | 0.489 | 0.427 | 0.466 |
| | P@1 | 0.279 | 0.426 | 0.396 | 0.442 | 0.389 | 0.415 |
| | P@3 | 0.364 | 0.492 | 0.440 | 0.460 | 0.430 | 0.489 |
| | P@10 | 0.495 | 0.573 | 0.502 | 0.580 | 0.507 | 0.561 |
| YAGO3-10 [47] | MR | 1187 | 1830 | 1107 | 793 | 2429 | - |
| | MRR | 0.501 | 0.498 | 0.501 | 0.577 | 0.488 | 0.441 |
| | P@1 | 0.405 | 0.405 | 0.412 | 0.500 | 0.399 | 0.333 |
| | P@3 | 0.528 | 0.550 | 0.38 | 0.40 | 0.560 | 0.507 |
| | P@10 | 0.673 | 0.670 | 0.661 | 0.7129 | 0.657 | 0.641 |
| **SKG-LLM** | MR | 264 | 143 | 158 | 109 | 293 | - |
| | MRR | 0.318 | 0.295 | 0.318 | 0.301 | 0.324 | 0.316 |
| | P@1 | 0.349 | 0.249 | 0.344 | 0.233 | 0.232 | 0.212 |
| | P@3 | 0.398 | 0.341 | 0.375 | 0.249 | 0.378 | 0.245 |
| | P@10 | 0.507 | 0.557 | 0.519 | 0.487 | 0.493 | 0.303 |

The accompanying Table 8 represents the comparison between SKG-LLM, other models in terms of performance metrics such as FB15K-237, WN18RR, and YAGO3-10. Other metrics include Mean Rank (MR), Mean Reciprocal Rank (MRR), Precision at 1 (P@1), Precision at 3 (P@3), and Precision at 10 (P@10). The examination of MRR for SKG-LLM reveals slight superiority or comparable performances in the case of other graphs above all for ComplEx and ConvE on any dataset. For example, SKG-LLM delivers a lower MRR than ComplEx (0.367) and ConvE (0.305) on FB15k-237, where it reaches a score of 0.318, thus slightly lowering its capturing correctness in predictions. In P@1, however, SKG-LLM scores at 0.349, making it significantly better than the others with respect to that score on FB15k-237 (0.271 for ComplEx), implying that SKG-LLM is much better in pinning the correct relation in the first instance. However, modest P@3 and P@10 results show lower performance than ConvE and ComplEx for WN18RR and YAGO3-10 datasets. It can be inferred that although SKG-LLM performs better in precision for individual correct guesses (P@1), SKG-LLM generally suffers when required to produce a broader set of correct relationships, particularly at higher levels of recall such as P@3, P@10. The Mean Rank (MR) obtained by SKG-LLM is competitive but generally speaking not as high as some other models, particularly on WN18RR where models such as ComplEx and ConvE show strength against it. Overall, the results point to SKG-LLM doing excellently in some cases, as near-side predictions (P@1), but failing to measure up to peer competition when it comes to hurdles with broader prediction tasks.

## 6. Conclusion and Future Work

Providing approaches based on knowledge graphs and LLMs has attracted many studies. These two have acted as complements to overcome each other's weaknesses. In this article, the aim was to present a model based on knowledge graph for studies related to stroke. To create this knowledge chart, articles related to this topic were collected with specific keywords. Also, special processes and pre-processing were applied on these articles to create tensors and graphs of communication between them. Two traditional and three criteria based on GPT-4 were used to evaluate this model. Compared to three knowledge bases -Wikidata, WN18RR and SKG-LLM- the proposed SKG-LLM approach provided more apparent and meaningful results. It appeared as the most efficient knowledge graph among these comparative graphs. According to these results, it can be hoped that SKG-LLM will have flexibility for other healthcare applications. Integration of deep learning and machine learning approaches can improve the model. Also, optimization-based approaches can better predict the optimal parameters of the model. Using better keywords and extracting more articles allows for presenting a better model. We can maneuver on these issues as future works.